# Interactive Policy Learning through
# Confidence-Based Autonomy

**Sonia Chernova**　　　　　　　　　　　　　　　　　SONIAC@CS.CMU.EDU
**Manuela Veloso**　　　　　　　　　　　　　　　　　VELOSO@CS.CMU.EDU
*Computer Science Dept.*
*Carnegie Mellon University*
*Pittsburgh, PA USA*

## Abstract

We present Confidence-Based Autonomy (CBA), an interactive algorithm for policy learning from demonstration. The CBA algorithm consists of two components which take advantage of the complimentary abilities of humans and computer agents. The first component, Confident Execution, enables the agent to identify states in which demonstration is required, to request a demonstration from the human teacher and to learn a policy based on the acquired data. The algorithm selects demonstrations based on a measure of action selection confidence, and our results show that using Confident Execution the agent requires fewer demonstrations to learn the policy than when demonstrations are selected by a human teacher. The second algorithmic component, Corrective Demonstration, enables the teacher to correct any mistakes made by the agent through additional demonstrations in order to improve the policy and future task performance. CBA and its individual components are compared and evaluated in a complex simulated driving domain. The complete CBA algorithm results in the best overall learning performance, successfully reproducing the behavior of the teacher while balancing the tradeoff between number of demonstrations and number of incorrect actions during learning.

## 1. Introduction

*Learning from demonstration* is a growing area of artificial intelligence research that explores techniques for programming autonomous agents by demonstrating the desired behavior or task. In demonstration-based approaches, a teacher, typically a human, shows the agent how to perform the task. The agent records the demonstrations as sequences of state-action pairs, from which it then learns a policy that reproduces the observed behavior. Many learning from demonstration approaches are inspired by the way humans and animals teach each other, aiming to provide an intuitive method to transfer human task knowledge to autonomous systems. Compared to exploration-based methods, demonstration learning often reduces the learning time and eliminates the frequently difficult task of defining a detailed reward function (Smart, 2002; Schaal, 1997).

In this article, we present an interactive demonstration learning algorithm, *Confidence-Based Autonomy* (CBA), which enables an agent to learn a policy through interaction with a human teacher. In this learning approach, the agent begins with no initial knowledge and learns a policy incrementally through demonstrations acquired as it practices the task. Each demonstration consists of a training point representing the correct action to be performed in a particular state. The agent's state is represented using an $n$-dimensional feature vector that can be composed of continuous or discrete values. The agent's actions are bound to a finite set $\mathcal{A}$ of action primitives, the basic actions that can be combined together to perform the overall task. Given a sequence of demonstrations $(s_i, a_i)$, with state $s_i$ and teacher-selected action $a_i \in \mathcal{A}$, the goal is for the agent to learn to imitate the teacher's behavior by generalizing from the demonstrations and learning a policy mapping from all possible states to actions in $\mathcal{A}$.

The method for gathering demonstrations is at the heart of all demonstration learning algorithms. CBA performs this function through two algorithmic components: Confident Execution, which enables the agent to select demonstrations in real time as it interacts with the environment using automatically calculated confidence thresholds, and Corrective Demonstration, which enables the teacher to improve the learned policy and correct mistakes through additional demonstrations. The complete Confidence-Based Autonomy algorithm provides a fast and intuitive method for policy learning, incorporating shared decision making between the learner and the teacher. In our experimental evaluation, we highlight the strengths of both learning components and compare learning performance of five different demonstration selection techniques. Our results indicate that in a complex domain, the Confident Execution algorithm reduces the number of demonstrations required to learn the task compared to demonstration selection performed by the human teacher. Additionally, we find that the teacher's ability to correct mistakes performed by the agent is critical for optimizing policy performance.

In Section 2, we discuss related work in learning from demonstration. We then present an overview of the complete Confidence-Based Autonomy learning algorithm in Section 3, followed by detailed descriptions of the Confident Execution and Corrective Demonstration components in Sections 4 and 5, respectively. In Section 6, we present an experimental evaluation of the complete algorithm and its components in a complex simulated driving domain. Section 7 presents a summary and discussion of possible extensions to this work.

## 2. Related Work

A wide variety of algorithms for policy learning from demonstration have been proposed within the machine learning and robotics communities. Within the context of reinforcement learning (Sutton & Barto, 1998), demonstration has been viewed as a source of reliable information that can be used to accelerate the learning process. A number of approaches for taking advantage of this information have been developed, such as deriving or modifying the reward function based on demonstrations (Thomaz & Breazeal, 2006; Abbeel & Ng, 2004; Papudesi, 2002; Atkeson & Schaal, 1997), and using the demonstration experiences to prime the agent's value function or model (Takahashi, Hikita, & Asada, 2004; Price & Boutilier, 2003; Smart, 2002; Schaal, 1997).

Demonstration has also been coupled with supervised learning algorithms for policy learning, including Locally Weighted Regression for low level skill acquisition (Grollman & Jenkins, 2007; Browning, Xu, & Veloso, 2004; Smart, 2002), Bayesian networks for high level behaviors (Lockerd & Breazeal, 2004; Inamura, Inaba, & Inoue, 1999), and the $k$-nearest neighbors algorithm for fast-paced games and robot navigation tasks (Saunders, Nehaniv, & Dautenhahn, 2006; Bentivegna, Ude, Atkeson, & Cheng, 2004). A recent survey covers these and other demonstration learning algorithms in detail (Argall, Chernova, Browning, & Veloso, 2009).

In addition to policy learning from demonstration, several areas of research have also explored algorithms for *demonstration selection*. Within machine learning research, active learning (Blum & Langley, 1997; Cohn, Atlas, & Ladner, 1994) enables a learner to query an expert and obtain labels for unlabeled training examples. Aimed at domains in which a large quantity of data is available but labeling is expensive, active learning directs the expert to label the more informative examples with the goal of minimizing the number of queries. In the context of reinforcement learning, the 'Ask For Help' framework enables an agent to request advice from other agents when it is "confused" about what action to take, an event characterized by relatively equal quality estimates for all possible actions in a given state (Clouse, 1996). Similarly motivated techniques have been used in robotics to identify situations in which a robot should request a demonstration from its teacher (Grollman & Jenkins, 2007; Lockerd & Breazeal, 2004; Nicolescu, 2003; Inamura et al., 1999). Most closely related to our work is the Dogged Learning algorithm (Grollman & Jenkins, 2007), a confidence-based learning approach for teaching low-level robotic skills. In this algorithm, the robot indicates to the teacher its certainty in performing various elements of the task. The teacher may then choose to provide additional demonstrations based on this feedback. While similarly motivated, our work differs from the Dogged Learning algorithm in a number of ways, most important of which are our use of classification instead of regression in policy learning, and our algorithm's ability to adjust the confidence threshold to the data instead of using a fixed value.

## 3. Confidence-Based Autonomy Overview

The Confence-Based Autonomy algorithm enables a human user to train a task policy through demonstration. The algorithm consists of two components:

- *Confident Execution (CE)*: an algorithm that enables the agent to learn a policy based on demonstrations obtained by regulating its autonomy and requesting help from the teacher. Demonstrations are selected based on automatically calculated classification confidence thresholds.

- *Corrective Demonstration (CD)*: an algorithm that enables the teacher to improve the learned policy by correcting mistakes made by the agent through supplementary demonstrations.

Figure 1 shows the interaction between these components. Using the Confident Execution algorithm, the agent selects states for demonstration in real time as it interacts with the environment, targeting states that are unfamiliar or in which the current policy action is uncertain. At each timestep, the algorithm evaluates the agent's current state and actively decides between autonomously executing the action selected by its policy and requesting an additional demonstration from the human teacher.

We assume the underlying model of the agent's task to be an MDP. The agent's policy is represented and learned using supervised learning based on training data acquired from the demonstrations. Confidence-Based Autonomy can be combined with any supervised learning algorithm that provides a measure of confidence in its classification. The policy is represented by classifier $\mathcal{C} : s \rightarrow (a, c, db)$, trained using state vectors $s_i$ as inputs, and actions $a_i$ as labels. For each classification query, the model returns the model-selected action $a \in \mathcal{A}$, action selection confidence $c$, and the decision boundary $db$ with the highest confidence for the query (e.g. Gaussian component for GMMs).

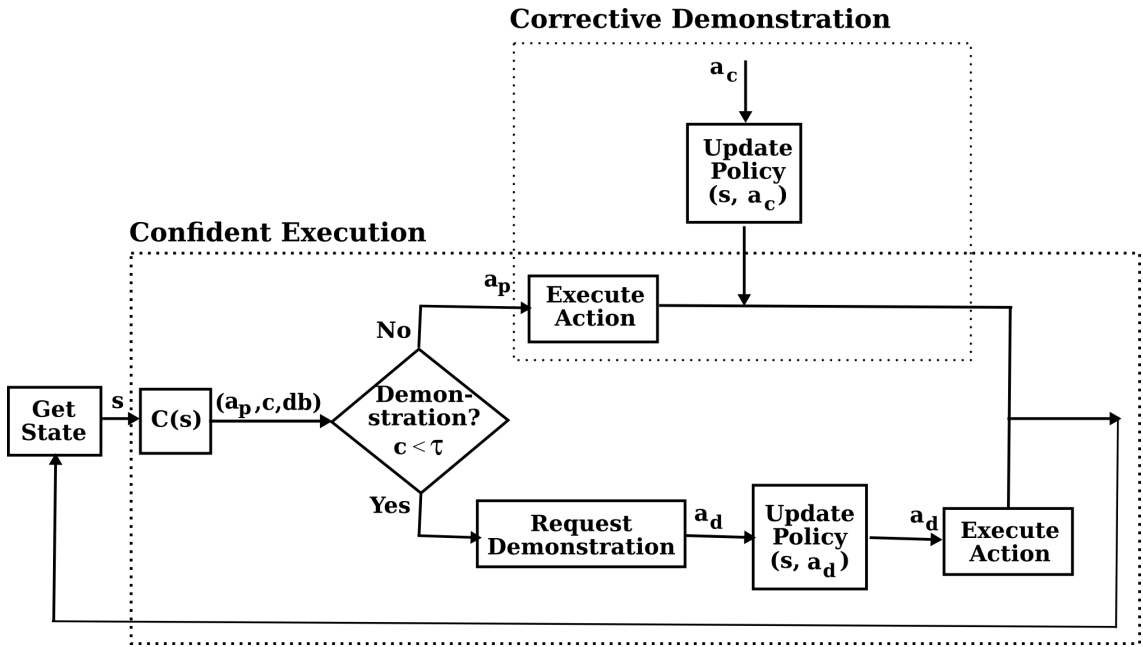

Figure 1: Confidence-Based Autonomy learning process.

To effectively select demonstrations, the learner must be able to autonomously identify situations in which a demonstration will provide useful information and improve the policy. Confident Execution selects between agent autonomy and a request for demonstration based on the measure of action-selection confidence $c$ returned by the classifier. Given the current state of the learner, the algorithm queries the policy to obtain its confidence in selecting an action for that state, and regulates its autonomy based on this confidence. The learner executes the returned action $a_p$ if confidence $c$ is above a threshold $\tau$, which is determined by the decision boundary of the classifier, $db$. Confidence below this threshold indicates that the agent is uncertain about which action to take, so it seeks help from the teacher in the form of a demonstration. Receiving an additional demonstration, $a_d$, in a low confidence situation improves the policy, leading to increased confidence, and therefore autonomy, in future similar states. As more training data becomes available, the quality of the policy improves and the autonomy of the agent increases until the entire task can be performed without help from the teacher. In Section 4 we compare two methods of using classification confidence to select states for demonstration.

Using the Confident Execution algorithm, the agent incrementally acquires demonstrations as it explores its environment. As it practices its task, the agent uses the policy it learned up to that point to make decisions between demonstration and autonomous execution. However, by relying on the policy before learning is complete, the algorithm is likely to make mistakes due to factors such as overgeneralization of the classifier or incomplete data in some area of the state space. To address this problem this article introduces the second algorithmic component, Corrective Demonstration, which allows the teacher to provide corrections for the agent's mistakes. Using this method, when an incorrect action is observed, the teacher provides an additional demonstration to the agent indicating which action *should* have been executed in its place. In addition to indicating that the wrong action was selected, this method also provides the algorithm with the correct action to perform in its place, $a_c$. The correction is therefore more informative than negative reinforcement or punishment techniques common in other algorithms, leading the agent to learn quickly from its mistakes.

Together, Confident Execution and Corrective Demonstration form an interactive learning algorithm in which the learner and human teacher play complimentary roles. The learner is able to identify states in which demonstration is required; in fact, our results show that the algorithm is able to do this better than the human teacher due to differences in perception and representation abilities. The teacher, on the other hand, possesses expert knowledge of the overall task, which is applied to performing demonstrations and spotting execution mistakes. This is a function the agent cannot perform on its own as it has not yet learned the desired behavior. In this way, Confidence-Based Autonomy takes advantage of the complimentary abilities of both human and agent. Sections 4 and 5 present the Confident Execution and Corrective Demonstration components in detail.

## 4. Confident Execution Algorithm

Confident Execution is an policy learning algorithm in which the agent must select demonstration examples, in real time, as it interacts with the environment. At each timestep, the algorithm uses thresholds to determine whether a demonstration of the correct action in the agent's current state will provide useful information and improve the agent's policy. If demonstration is required, the agent requests help from the teacher, and updates its policy based on the resulting action label. Otherwise the agent continues to perform its task autonomously based on its policy.

There are two distinct situations in which the agent requires help from the teacher, *unfamiliar* states and *ambiguous* states. An unfamiliar state occurs when the agent encounters a situation that is significantly different from any previously demonstrated state, as represented by the outlying points in Figure 2. While we do not want to demonstrate every possible state, and therefore need our model to generalize, we would like to prevent over-generalization to truly different states.

Ambiguous states occur when the agent is unable to select between multiple actions with certainty. This situation can result when demonstrations of different actions from similar states make accurate classification impossible, as in the region of overlapping data classes in Figure 2. In these cases, additional demonstrations may help to disambiguate the situation.

The goal of the Confident Execution algorithm is to divide the state space into regions of high confidence (autonomous execution) and low confidence (demonstration) such that unfamiliar and ambiguous regions fall into the low confidence areas. Given a world state, two evaluation criteria are used to select between demonstration and autonomy:

- *Nearest Neighbor distance*: Given $d = NearestNeighbor$(s), the distance from the current state to the nearest (most similar) training datapoint, the agent may act autonomously if $d$ is below the distance threshold $\tau_{dist}$.

- *Classification confidence*: Given $c$, the classification confidence of the current state, the agent may act autonomously if the value of $c$ is above the confidence threshold $\tau_{conf}$.

The methods for calculating thresholds $\tau_{dist}$ and $\tau_{conf}$ are presented in Sections 4.1 and 4.2. In this section, we continue the discussion of the Confident Execution algorithm assuming that these values are given.

Algorithm 1 presents the details of the Confident Execution algorithm. We assume no preexisting knowledge about the task, and initialize the algorithm with an empty set of training points $T$. Since a classifier is not initially available, threshold $\tau_{conf}$ is initialized to infinity to ensure that the agent is controlled through demonstration during the initial learning stage. Distance threshold $\tau_{dist}$ is initialized to 0.

The main learning algorithm consists of a loop (lines 4-20), each iteration of which represents a single timestep. The behavior of the algorithm is determined by whether the agent is currently executing an action. If an action is in progress, the algorithm performs no additional computation during this timestep (line 20). Once an action is complete, the algorithm evaluates its state to determine the next action to perform (lines 6-18).

Evaluation begins by obtaining the agent's current state in the environment (line 6). This information is then used to calculate the nearest neighbor distance and to query the learned classifier $\mathcal{C}$ to obtain policy action $a_p$ and confidence $c$. These values are then compared to the confidence and distance thresholds to decide between demonstration and autonomy (line 9). If similar states have previously been observed, and the learned model is confident in its selection, the algorithm finishes the timestep by initiating the autonomous

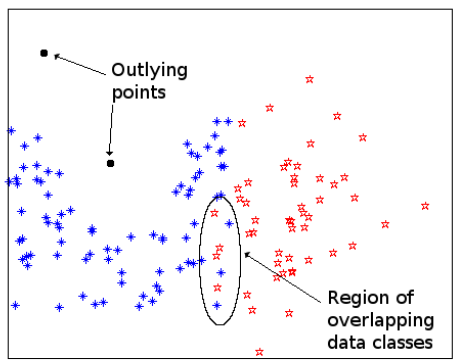

Figure 2: Outlying points and regions of overlapping data classes represent unfamiliar and ambiguous state regions, respectively.

---

**Algorithm 1** Confident Execution Algorithm

---

1: $T \leftarrow \{\}$
2: $\tau_{conf} \leftarrow \inf$
3: $\tau_{dist} \leftarrow 0$
4: **while true do**
5:   **if** `actionComplete` **then**
6:     $s \leftarrow \text{GetSensorData}()$
7:     $d = \text{NearestNeighbor}(s)$
8:     $(a_p, c, db) \leftarrow \mathcal{C}(s)$
9:     **if** $c > \tau_{conf}$ **and** $d < \tau_{dist}$ **then**
10:       $\text{ExecuteAction}(a_p)$
11:     **else**
12:       $\text{RequestDemonstration}()$
13:       $a_d \leftarrow \text{GetTeacherAction}()$
14:       **if** $a_d \neq NULL$ **then**
15:         $T \leftarrow T \cup \{(s, a_d)\}$
16:         $\mathcal{C} \leftarrow \text{UpdateClassifier}(T)$
17:         $(\tau_{conf}, \tau_{dist}) \leftarrow \text{UpdateThresholds}()$
18:         $\text{ExecuteAction}(a_d)$
19:   **else**
20:     `//do nothing`

---

execution of the policy selected action $a_p$ (line 10). Otherwise it initiates a request for teacher demonstration (lines 12-18).

The agent requests a demonstration by pausing and indicating to the teacher that a demonstration is required. Note that we assume the domain allows the agent to pause execution. Following a demonstration request, the algorithm checks whether a demonstration has been performed (lines 13-14). If the teacher's response is available, a new training datapoint consisting of the current state and the corresponding demonstrated action $a_d$ is added to the training set (line 15). The model classifier is then retrained, and the threshold values updated, before executing the teacher selected action (lines 16-18).

If the teacher's response is not immediately available, the timestep terminates and the whole process is repeated at the next iteration. The agent again senses its state, performs the threshold comparison and checks for a demonstration. This non-blocking mechanism enables the agent to wait for a demonstration from the teacher without losing awareness of its surroundings. In cases where the agent's environment is dynamic, maintaining up to date information is important as the state may change in the time between the initial request and the demonstration. Associating the action label with the agent's most recent state, the one the teacher is most likely responding to, is therefore critical to learning an accurate model. Additionally, changes in the environment can result in the agent attaining a high confidence state without any actions of its own. In these cases, autonomous execution of the task is automatically resumed. In summary, once a demonstration request is made, no further actions are taken by the agent until either a demonstration is received from the teacher, or changes in the environment result in a high confidence state.

Using this approach, Confident Execution enables the agent to incrementally acquire demonstrations representing the desired behavior. As more datapoints are acquired, fewer states distant from the training data are encountered, the performance and classification confidence improve, and the autonomy of the agent increases. Task learning is complete once the agent is able to repeatedly perform the desired behavior without requesting demonstrations. In the following sections we present the methods for calculating the distance and confidence thresholds.

## 4.1 Distance Threshold

The purpose of the distance threshold is to evaluate the similarity between the agent's current state and previous demonstrations. Our evaluation metric uses the nearest neighbor distance, defined as a the Euclidian distance between a query and the closest point in the dataset. For each agent state query, we obtain its nearest neighbor distance representing the most similar previously demonstrated state. This value is then compared to the distance threshold $\tau_{dist}$.

The value of the distance threshold $\tau_{dist}$ is calculated as a function of the average nearest neighbor distance across the dataset of demonstrations. Evaluating the average similarity between states provides the algorithm with a domain-independent method for detecting outliers, points unusually far from previously encountered states. For trials in this article, the value of $\tau_{dist}$ was set to three times the average nearest neighbor distance across the dataset.

An alternate method for detecting outliers would be to use classification confidence and request demonstrations in low confidence states. However, situations can arise in which confidence is not directly correlated with state similarity. For example, for many classifiers a set of datapoints encircling an empty region, similar to the shape of a donut, would result in the highest classification confidence being associated with the empty center region far from previous demonstrations. Distance provides a reliable prediction of similarity, even in these cases.

## 4.2 Confidence Threshold

The confidence threshold is used to select regions of uncertainty in which points from multiple classes overlap. From the agent's perspective, points in these regions represent demonstrations of two distinct actions from states that appear similar, and are difficult to distinguish based on the sensor data. This problem frequently arises in demonstration learning for a number of reasons, such as the teacher's inability to demonstrate the task consistently, noise in the sensor readings, or an inconsistency between the agent's and teacher's sensing abilities. We would like to set the confidence threshold to a value that prevents either model from classifying the overlapping region with high confidence[1]. In the following section we will discuss the use and limitations of a single fixed threshold value. We then present an algorithm for using multiple adjustable thresholds in Section 4.2.2.

---

1. See Section 7.2 for further discussion of these data regions.

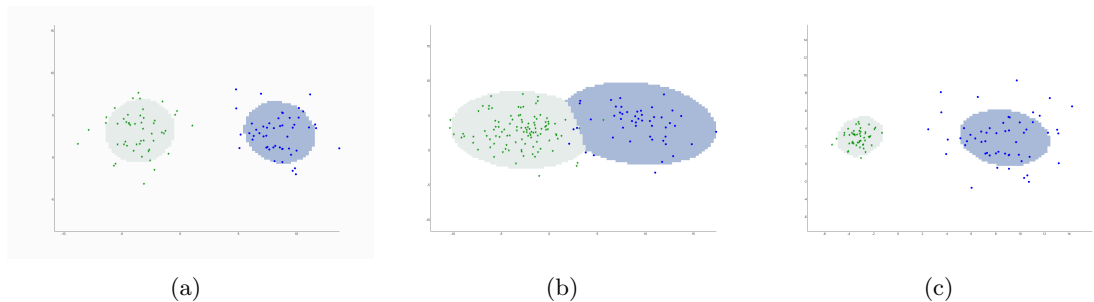

Figure 3: Examples of fixed threshold failure cases: (a) Fully separable data classes with an overly conservative threshold value (b) Overlapping data classes with an overly general threshold value (c) Data classes with different distributions and common threshold value

### 4.2.1 Single Fixed Threshold

A single, fixed confidence threshold value provides a simple mechanism to approximate the high confidence regions of the state space. Previous algorithms utilizing a classification confidence threshold for behavior arbitration have all used a manually-selected single threshold value (Inamura et al., 1999; Lockerd & Breazeal, 2004; Grollman & Jenkins, 2007). However, choosing an appropriate value can be difficult for a constantly changing dataset and model. Figure 3 presents examples of three frequently encountered problems.

Figure 3(a) presents a case in which two action classes are distinct and fully separable. A model trained on this dataset is able to classify the points with complete accuracy, without misclassifications. However, the current threshold value classifies only 72% of the points with high confidence, marking the remaining 28% of the points as uncertain. In this case, a lower threshold value would be preferred that would allow the model to generalize more freely. The resulting larger high confidence region would reduce the number of redundant demonstrations without increasing the classification error rate of either data class.

Figure 3(b) presents an example of the opposite case, in which a stricter threshold value would be preferred. In this example the data classes overlap, resulting in a middle region in which points cannot be classified with high accuracy. A higher threshold value would prevent the classification of points in this region into either data class, initiating instead a request for demonstration that would allow the teacher to disambiguate the situation.

Figure 3(c) presents a case in which the datapoints of the two data classes have very different distributions. While the fixed threshold value is appropriate for the left class, 42% of the points in the right class are labeled as low confidence.

Classification of complex multi-class data depends upon multiple decision boundaries. Using the same value for all decision boundaries can exacerbate the problems highlighted above, as a single value often cannot be found that constrains model classification in some areas while allowing generalization in others. The resulting effect is that the agent requests too many demonstrations about things it already knows, and too few demonstrations about unlearned behavior. To address this problem, we present an algorithm for calculating a unique threshold value for each decision boundary.

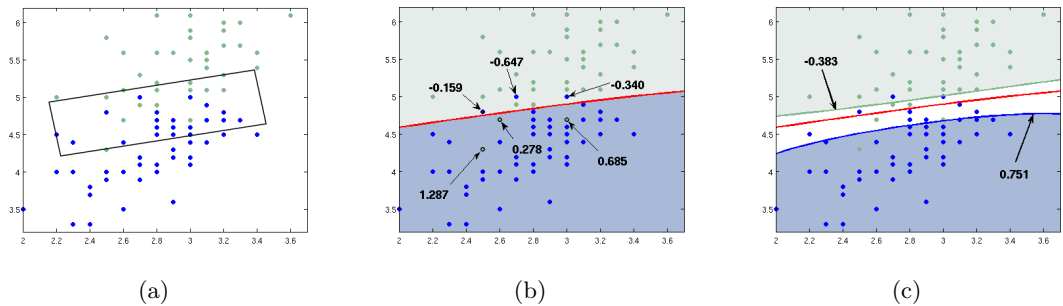

Figure 4: Autonomy threshold calculation: (a) Example dataset, with highlighted overlapping region (b) Learned decision boundary, misclassified points marked with confidence values (c) Learned threshold values for each data class, a low confidence region containing most of the overlapping points remains in the center.

### 4.2.2 Multiple Adjustable Thresholds

In this section, we contribute an algorithm for calculating a confidence threshold for each decision boundary, customized to its unique distribution of points. In our analysis, we assume that we are able to query the classifier and obtain a confidence score representing the likelihood that a particular input belongs within a specified decision boundary.

The algorithm begins by dividing the dataset into a training and test set and training the classifier $\mathcal{C}$. The resulting learned model is used to classify the withheld test set, for which the correct action labels are known. The algorithm then calculates a unique confidence threshold for each decision boundary based on the confidence scores of misclassified points. Given the confidence scores of a set of points mistakenly classified by a decision boundary, *we assume that future classifications with confidences at or below these values are likely to be misclassifications as well.* The threshold is therefore calculated as a function of these confidence scores.

Specifically, we define a classified point as the tuple $(o, a, a_m, c)$, where $o$ is the original observation, $a$ is the demonstrated action label, $a_m$ is the model-selected action, and $c$ is the model action confidence. Let $M_i = \{(o, a_i, a_m, c) | a_m \neq a_i\}$ be the set of all points mistakenly classified by decision boundary $i$. The confidence threshold value is set to the average classification confidence of the misclassified points: $\tau_{conf_i} = \frac{\sum^{M_i} c}{|M_i|}$. We take the average to avoid overfitting to noisy data. Other values, based on the maximum or standard deviation, can be used if a more conservative estimate is required. A threshold value of 0 indicates that no misclassifications occurred and the model is able to generalize freely.

Figure 4 presents an example of the threshold calculation process. Figure 4(a) presents a small sample dataset, the rectangular box in the figure highlights a region of the state space in which points from both classes overlap. Figure 4(b) shows the learned decision boundary (in this case a SVM) separating our two data classes. Six misclassified points are marked with the (mis-)classification confidences returned by the model. Misclassified points on each side of the decision boundary will be used to calculate the respective confidence thresholds. Figure 4(c) shows the confidence threshold lines and values based on the above calculations. The resulting low confidence region in the middle of the image captures most of the noisy datapoints.

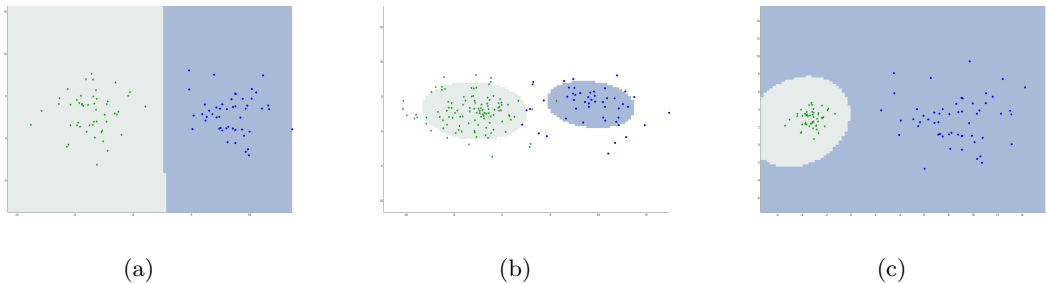

Figure 5: Multiple adjustable thresholds applied to the failure cases shown in Figure 3.

Given this multi-threshold approach, classification of new points is performed by first selecting the action class with the highest confidence for the query. The comparison on line 9 of Algorithm 1 is then performed using the threshold of the decision boundary with the highest confidence for the query. Using this method, the threshold value of the most likely decision boundary to represent the point is used to decide between demonstration and autonomy.

Figure 5 shows how the example failure cases discussed in Section 4.2.1 are addressed by the multi-thresholded approach. Customizing the threshold value to each unique data distribution enables the algorithm to correctly classify 100% of the points in Figures 5(a) and (c). Since there are no misclassifications, the model generalizes freely in these examples. For the dataset in Figure 5(b), in which perfect classification is not possible, the confidence thresholds are set such that the overlapping region falls into a low confidence area. This example uses a Gaussian mixture model, in which the elliptical confidence gradient around the mean results in a large low confidence area even far from the overlapping region. Other classification methods, such as Support Vector Machines, do not have this drawback.

The presented multi-threshold approach is algorithm independent, and Figure 6 presents classification results of four different classification methods: Gaussian mixture models, random forests (RF), Support Vector Machine with a quadratic kernel, and SVM with a radial basis function (RBF) kernel. The table below summarizes the classification performance of each algorithm and lists the threshold values for each of the models.

| Algorithm | Correct-Misclas.-Unclass. | Thresholds |
|---|---|---|
| GMM | $98.6\% - 0.4\% - 1.0\%$ | (0, 0, 0.012) |
| RF | $99.1\% - 0.1\% - 0.8\%$ | (0.14, -0.355) |
| SVM quad. | $98.5\% - 0.1\% - 1.4\%$ | (335.33, -68.77) |
| SVM RBF | $98.9\% - 0.1\% - 1.0\%$ | (0.825, -0.268) |

Table 1: Classifier comparison.

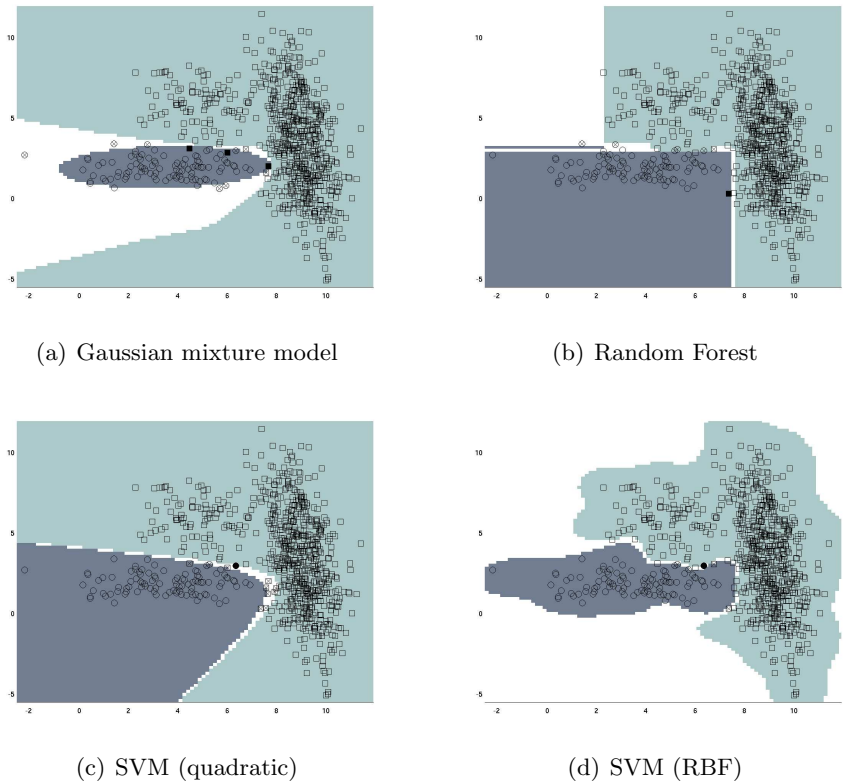

(a) Gaussian mixture model

(b) Random Forest

(c) SVM (quadratic)

(d) SVM (RBF)

Figure 6: Classification of dataset into high and low confidence regions using different classification methods.

## 5. Corrective Teacher Demonstration

The presented Confident Execution algorithm enables the agent to identify unfamiliar and ambiguous states and prevents autonomous execution in these situations. However, states in which an incorrect action is selected with high confidence for autonomous execution still occur, typically due to over-generalization of the classifier. In this article we present the Corrective Demonstration algorithm which, coupled with Confident Execution, enables the teacher to correct mistakes made by the agent. Algorithm 2 combines Corrective Demonstration (lines denoted by $\star$) with Confident Execution and presents the complete Confidence-Based Autonomy algorithm.

The Corrective Demonstration technique comes into play each time the agent executes an autonomous action. As an action is selected for autonomous execution, the algorithm records the agent's state that led to this decision and saves this value within the variable $s_c$ (line 11). During the execution of an autonomously selected action, the algorithm checks for a teacher demonstration at every timestep (lines 22-23). If a corrective demonstration is made, a new training datapoint consisting of the recorded demonstration state $s_c$ and the corrective action $a_c$ is added to the training set (line 24). The classifier and thresholds are then retrained using the new information.

---

**Algorithm 2** Confidence-Based Autonomy algorithm: Confident Execution and Corrective Demonstration

---

1: $T \leftarrow \{\}$
2: $\tau_{conf} \leftarrow \inf$
3: $\tau_{dist} \leftarrow 0$
4: **while true do**
5:    $s \leftarrow$ GetSensorData()
6:    **if** `actionComplete` **then**
7:       $(a_p, c, db) \leftarrow \mathcal{C}(s)$
8:       $d =$ NearestNeighbor(s)
9:       **if** $c > \tau_{conf}$ **and** $d < \tau_{dist}$ **then**
10:          ExecuteAction($a_p$)
11:          $s_c \leftarrow s$                                      ★
12:       **else**
13:          RequestDemonstration()
14:          $a_d \leftarrow$ GetTeacherAction()
15:          **if** $a_d \neq NULL$ **then**
16:             $T \leftarrow T \cup \{(s, a_d)\}$
17:             $\mathcal{C} \leftarrow$ UpdateClassifier($T$)
18:             $(\tau_{conf}, \tau_{dist}) \leftarrow$ UpdateThresholds()
19:             ExecuteAction($a_d$)
20:    **else**
21:       **if** $autonomousAction$ **then**                   ★
22:          $a_c \leftarrow$ GetTeacherAction()               ★
23:          **if** $a_c \neq NULL$ **then**              ★
24:             $T \leftarrow T \cup \{(s_c, a_c)\}$         ★
25:             $\mathcal{C} \leftarrow$ UpdateClassifier($T$)      ★
26:             $(\tau_{conf}, \tau_{dist}) \leftarrow$ UpdateThresholds()    ★

---

Using this algorithm, the teacher observes the autonomous execution of the agent and corrects any incorrect actions. Unlike our previous demonstration technique in which the agent was given the next action to perform, the correction is performed with relation to the agent's *previous* state at which the mistake was made. For example, when observing a driving agent approaching too close behind another car, the teacher is able to indicate that instead of continuing to drive forward, the agent should have been merging into the passing lane. In this way, in addition to indicating that the wrong action was performed, Corrective Demonstration also provides the algorithm with the action that should have been performed in its place. This technique is more effective than negative reinforcement, or punishment, techniques common in other algorithms, leading the agent to learn quickly from its mistakes.

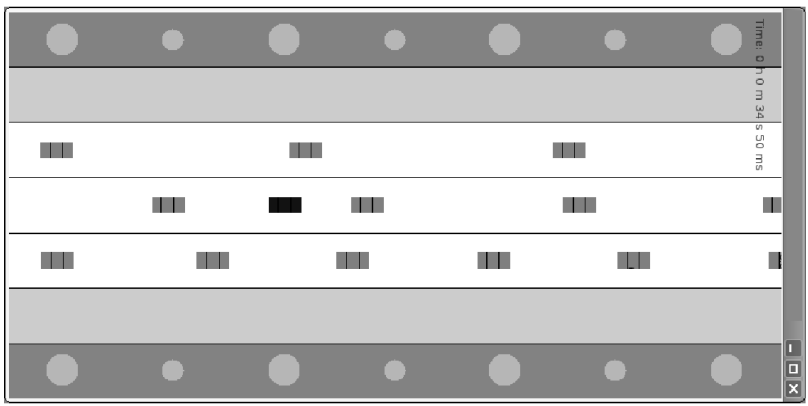

Figure 7: Screenshot of the driving simulator. The agent, the black car currently in the center lane, drives at a fixed speed and must navigate around other cars to avoid collisions. The road consists of five lanes: three traffic lanes and two shoulder lanes.

## 6. Evaluation and Comparison

In this section we present an evaluation and comparison of the complete Confidence-Based Autonomy algorithm and its components in simulated car driving domain (Abbeel & Ng, 2004), shown in Figure 7.

### 6.1 Domain Description

In the driving domain, the agent represents a car driving on a busy highway. The learner's car travels at a fixed speed of 60 mph, while all other cars move in their lanes at predetermined speeds between 20 and 40 mph. The road has three normal lanes and a shoulder lane on both sides; the agent is allowed to drive on the shoulder to pass other cars, but cannot go further off-road. Since the learner cannot change its speed, it must navigate between other cars and use the shoulder lanes to avoid collision. The agent is limited to three actions: remaining in the current lane, or shifting one lane to the left or right of the current position ($\mathcal{A} = \{forward, left, right\}$). The teacher demonstrates the task through a keyboard interface. The simulator has a framerate of 5 fps and is paused during demonstration requests.

The agent's state is represented by: $s = \{l, d_l, d_c, d_r\}$. State feature $l$ is a discrete value symbolizing the agent's current lane number. The remaining three features, denoted by the letter $d$, represent the distance to the nearest car in each of the three driving lanes (left, center and right). The distance features are continuously valued in the [-25,25] range; note that the nearest car in a lane can be behind the agent. Distance measurements are corrupted by noise to create a more complex testing environment. The agent's policy is relearned each time 10 new demonstrations are acquired.

The driving domain presents a varied and challenging environment; if car distances were to be discretized by rounding to the nearest integer value, the domain would contain over 600,000 possible states. Due to the complexity of the domain, the agent requires a large number of demonstrations to initialize the classifier, resulting in nearly constant demonstration requests early in the training process. To simplify the task of the teacher, we add a short 300 datapoint, or approximately 60 second, non-interactive driving demonstration session to initialize the learning process. While this learning stage is not required, it simplifies the task of the teacher for whom continuous demonstration is preferred over frequent pauses for demonstration requests.

The performance of each learning algorithm was evaluated each time 100 new demonstrations were acquired. For each evaluation, the agent drove for 1000 timesteps over a road segment with a fixed and consistent traffic pattern. This road segment was not used for training, instead each algorithm was trained using a randomly generated car traffic pattern.

Since the algorithm aims to imitate the behavior of the expert, no 'true' reward function exists to evaluate the performance of a given policy. We present two domain-specific evaluation metrics that capture the key characteristics of the driving task. Our first evaluation metric is the agent's *lane preference*, or the proportion of the time the agent spends in each lane over the course of a trial. This metric provides an estimate of the similarity in driving styles. Since the demonstrated behavior attempts to navigate the domain without collisions, our second evaluation metric is the *number of collisions* caused by the agent. Collisions are measured as the percentage of the total timesteps that the agent spends in contact with another car. Always driving straight and colliding with every car in the middle lane results in a 30% collision rate.

## 6.2 Experimental Results

We present the performance evaluation and comparison of the following demonstration selection techniques:

- $TG$ – Teacher-guided, all demonstrations selected by the teacher without any confidence feedback from the algorithm and without the ability to perform retroactive corrections

- $CE_S$ – Confident Execution, all demonstrations selected by the agent using a single fixed confidence threshold

- $CE_M$ – Confident Execution, all demonstrations selected by the agent using multiple adjustable confidence thresholds

- $CD$ – Corrective Demonstration, all demonstrations selected by the teacher and performed as corrections in response to mistakes made by the agent

- $CBA$ – The complete Confidence-Based Autonomy algorithm combining Confident Execution using multiple adjustable confidence thresholds with Corrective Demonstration

For each demonstration selection method, the underlying policy of the agent was learned using multiple Gaussian mixture models, one for each action class (Chernova & Veloso, 2007). Videos of the driving task are available at *www.cs.cmu.edu/∼soniac*.

Figure 8 presents performance results of the five algorithms with respect to the above defined lane preference and collision metrics. We describe and discuss all elements of the

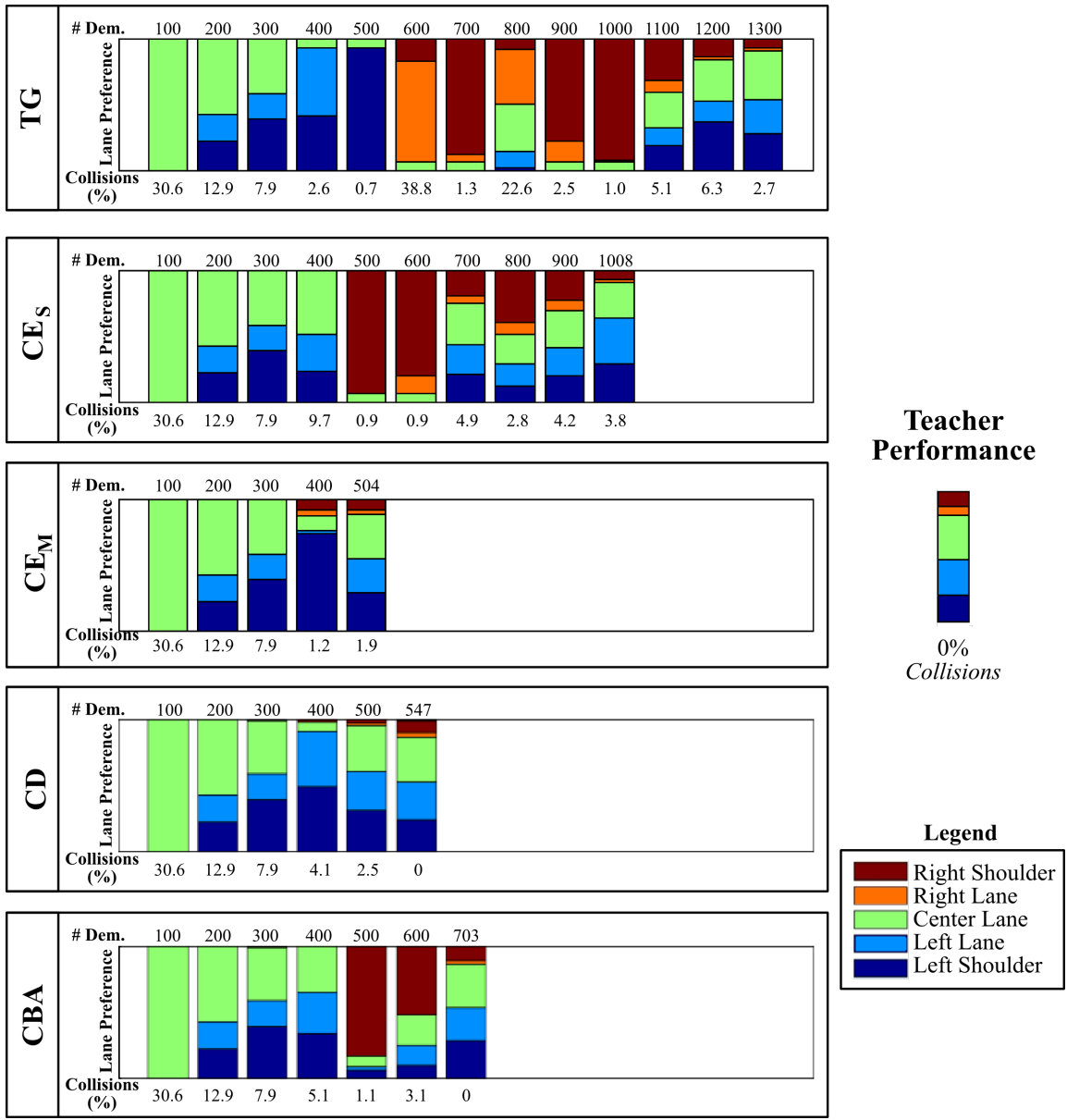

Figure 8: Evaluation of the agent's driving performance at 100-demonstration intervals for each of the five demonstration selection methods. The bar graphs indicate the percentage of time the agent spent in each road lane. Values under each bar indicate the percentage of collision timesteps accrued over the evaluation trial. The teacher performance bar on the right of the figure shows the teacher's driving lane preference and collision rate over the evaluation road segment. The goal is for each algorithm to achieve performance similar to that of the teacher.

figure in detail in the following sections. For each evaluation, the figure presents a bar representing a composite graph showing the percentage of time spent by the agent in each lane. The value above the bar indicates the number of demonstrations upon which the evaluated policy is based. The value below the bar indicates the percentage of incurred collisions during the evaluation.

The bar on the right of the figure shows the performance of the teacher over the evaluation road segment. This evaluation indicates that the teacher prefers to drive in the center and left lanes, followed in preference by the left shoulder, right shoulder and right lane. The teacher also successfully avoids all collisions, resulting in a collision rate of 0%. The goal of the learning algorithm is to achieve a driving lane pattern similar to that of the teacher and also without collisions. Note that, as described in the previous section, policy learning was initialized with the same 300-demonstration dataset for all algorithms. This initialization results in identical performance across all algorithms for this initial learning segment.

### 6.2.1 *TG* Demonstration Selection

The top row in Figure 8 summarizes the performance of the teacher-guided demonstration selection approach. In this approach, the teacher performed training by alternating between observing the performance of the agent and selecting demonstrations that, in her opinion, would improve driving performance. The teacher selected all training examples without receiving feedback about action selection confidence, and without the ability to provide corrective demonstrations for incorrect actions that were already executed by the agent. Instead, the teacher was required to *anticipate* what data would improve the policy. The training process was terminated once the teacher saw no further improvement in agent performance.

Figure 8 shows the results of the agent's performance evaluations at 100-demonstration intervals throughout the learning process. The similarity in the driving lane preference of the agent improves slowly over the course of the learning, with significant fluctuations. For example, after 500 demonstrations, the agent's preference is to drive on the empty left shoulder, thereby incurring few collisions. One hundred demonstrations later, the policy has shifted to prefer the center lane. However, the agent has not yet learned to avoid other cars, resulting in a 38.8% collision rate. The policy stabilizes after approximately 1100 demonstrations, representing a driving style similar to that of the teacher, with a small number of collisions. Without confidence feedback from the agent, it is difficult for the teacher to select an exact termination point for the learning. Training continued until, after 1300 demonstrations, the learner's policy showed little improvement. The final policy resulted in a lane preference very similar to that of the expert, but with a 2.7% collision rate.

### 6.2.2 *CE_S* Demonstration Selection

The second row in Figure 8 presents the results of the Confident Execution algorithm with a single autonomy threshold. In this demonstration selection approach, all demonstrations were selected by the agent and learning terminated once the agent stopped requesting demonstrations and performed all actions autonomously. The autonomy threshold value was selected by hand and evaluated in multiple performance trials. Results of the best fixed threshold are presented.

Compared to the teacher-guided approach, the policy learned using the $CE_S$ algorithm stabilizes quickly, achieving performance similar to the teacher's after only 700 demonstrations. The number of collisions is again low but persistent, even as the agent gains full confidence and stops requesting demonstrations after 1008 demonstrations. The final lane preference was again similar to that of the expert, with a collision rate of 3.8%.

### 6.2.3 $CE_M$ Demonstration Selection

The third row in Figure 8 presents the results of the Confident Execution algorithm with multiple autonomy thresholds, which were calculated using the algorithm presented in Section 4.2.2. Of all the demonstration selection methods, $CE_M$ required the fewest number of demonstrations to learn the task, completing learning after only 504 demonstrations. This result indicates that the use of multiple adjustable thresholds successfully focuses demonstration selection on informative areas of the state space while greatly reducing the number of redundant demonstrations. Throughout the learning process, the number of Gaussian components within the model varied between 9 and 41. This large variation highlights the importance of automating the threshold calculation process, since hand-selecting individual thresholds for each component would be impractical. The lane preference of the final policy was again similar to that of the expert. However, the agent still maintained a small collision rate of 1.9%.

### 6.2.4 $CD$ Demonstration Selection

The evaluation of the first three algorithms highlights the difficulty of the driving problem. Each of the approaches was able to select demonstrations that resulted in a policy that mimics the overall driving style of the teacher. However, all of the policies resulted in a small number of collisions, which typically occurred when the agent merged too close to another vehicle and touched its bumper. Such mistakes are difficult to correct using the techniques evaluated so far. Even within the teacher guided demonstration selection method, in which the human teacher has full control of the demonstration training data, by the time the collision has been observed the incorrect decision had already been made by the algorithm. Instead, retroactive demonstration is required to correct already made mistakes, as in the Corrective Demonstration algorithm.

In the fourth row of Figure 8 we present our evaluation of demonstration selection using only the Corrective Demonstration algorithm. In this approach, all demonstrations were selected by the teacher as corrections in response to mistakes made by the agent. Behavior corrected by the teacher included collisions, as well as incorrect lane preference (e.g. always driving on the shoulder) and rapid oscillations between lanes. To enable the teacher to accurately perform corrections, the simulation was slowed from 5 to 2 frames per second. Learning was terminated once the agent required no further corrections. As shown in Figure 8, the complete training process using Corrective Demonstration took 547 demonstrations, achieving a final policy that correctly imitates the teacher's driving style with a 0% collision rate. In the following section, we discuss how this performance compares to the complete CBA algorithm.

### 6.2.5 *CBA* Demonstration Selection

The final row in Figure 8 presents the evaluation of the complete Confidence-Based Autonomy algorithm, which combines $CE_M$ with $CD$. Using this approach, learning is complete once the agent no longer requests demonstrations *and* is able to perform the driving task without collisions. Using CBA the agent required a total of 703 demonstrations to learn the task, successfully learning to navigate the highway without collisions.

We analyze the impact of the two CBA learning components by comparing the number and distribution of demonstrations acquired by each algorithm during the learning process. In this section we refer to the learning components of CBA as CBA-CE and CBA-CD to differentiate from the algorithm evaluations presented in previous sections. Note that the behavior of the Confident Execution component is dependent upon the method used to set the autonomy thresholds. In this evaluation we use multiple adjustable thresholds calculated as the average value of misclassified points.

In Figure 9(a), each datapoint along the x-axis represents the number of demonstrations requested using CBA-CE (top) and initiated by the teacher using CBA-CD (bottom) during a 100-timestep interval, or approximately 40 seconds of simulator runtime (excluding pauses for demonstration requests). Since the first three 100-demonstration timesteps consist entirely of non-interactive demonstration, the values for these timesteps are 100 and, due to scaling, exceed the bounds of the graph. Figure 9(b) shows how the cumulative number of demonstrations for each component, and in total, grows with respect to training time. The complete training process lasts approximately an hour and a half.

Analysis of these graphs shows that most demonstrations occur early in the training process. Importantly, Confident Execution accounts for 83% of the total number of demon-strations, indicating that the agent guides most of the learning. Most of these demonstration requests occur during the first few minutes of training when the agent encounters many novel states and the classification confidence remains low. The agent requires few corrections during this stage because many mistakes are prevented by requesting a demonstration instead of performing a low confidence action. Corrective Demonstration plays its greatest role towards the end of training process, where it accounts for 73% of the final 100 demonstrations. At this stage in the learning the agent's action selection confidence is high enough that it rarely asks for demonstrations. Its policy already closely imitates the teacher's driving style but a small number of collisions remain. Corrective Demonstration enables the teacher to fine-tune the policy and eliminate all collisions. This result highlights the importance of Corrective Demonstration, whether alone or in conjunction with another selection technique, for optimizing policy performance.

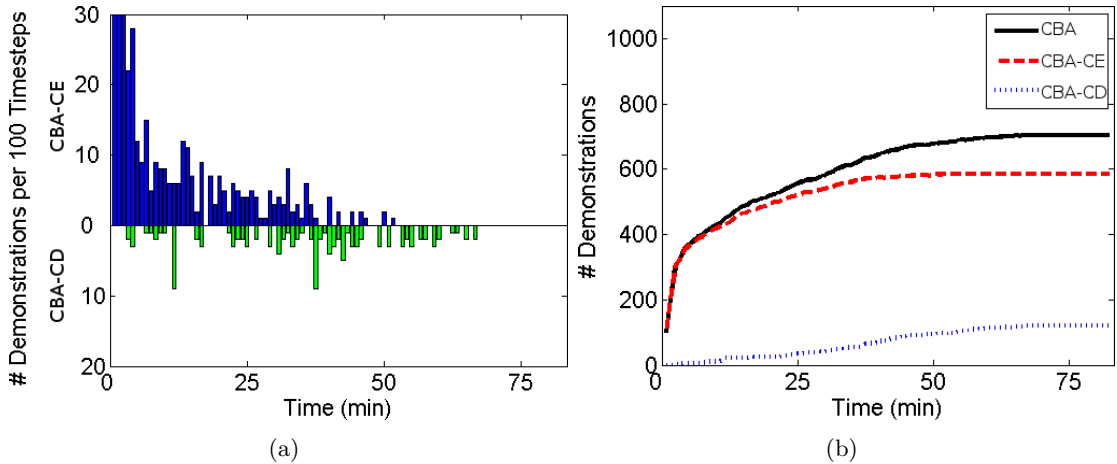

(a)                                              (b)

Figure 9: (a) Timeline showing how the number of demonstrations initiated by the agent through Confident Execution (top) and initiated by the teacher through Corrective Demonstrations (bottom) changes over the course of the training. (b) The cumulative number of demonstrations acquired by each component, and in total, over time.

While CBA achieves similar final performance compared to the CD algorithm evaluated in the previous section, it requires approximately 150 additional demonstrations to learn this policy. The additional demonstrations can be attributed to Confident Execution demonstration requests that served to increase the classification confidence but did not change the outcome of the agent's action. Viewed another way, these datapoints correspond to states in which the agent would have performed the correct action even if it had not asked for a demonstration. From this result it appears that allowing the agent to make mistakes and correcting them after the fact, as done in the CD evaluation, may be the best demonstration selection approach with respect to the performance metrics defined above and the overall number of demonstrations.

However, eliminating the ability to request demonstrations and utilizing only retroactive correction has several drawbacks, namely requiring constant and full attention from the teacher, and, most importantly, requiring our agent to make many mistakes before it learns the correct policy. By comparison, the CBA algorithm enables the agent to request demonstrations in low confidence states, thereby avoiding many incorrect actions. Our original lane preference and collision metrics do not take this difference into account as they focus only on the final policy performance of the agent.

To evaluate the difference between these algorithms, we additionally examine the number of collisions each agent incurs over the course of the learning. Using the CD algorithm, the agent incurs 48% more collisions (278 vs. 188) during training than by using CBA. Therefore, by allowing the agent to request demonstrations in low-confidence states, the CBA algorithm requires a slightly greater number of demonstrations while greatly reducing the number of incorrect actions performed during learning. The reduction in the number of action errors is significant due to its importance for many learning domains, especially robotic applications in which such errors may pose dangers to the system.

In summary, our evaluation has shown that the ability to retroactively *correct mistakes* is crucial to optimizing the policy and eliminating all collisions. The best performance was achieved by the Corrective Demonstration and Confidence-Based Autonomy methods, with CD requiring fewer demonstrations but incurring a greater number of collisions during training. The choice between CD and CBA can therefore be viewed as a tradeoff between the number of demonstrations and the frequency of undesired actions during training. In fact, CD is a special case of CBA in which the autonomy threshold is set to classify all points with high confidence. Adjusting the selectiveness of the CBA autonomy thresholds could, therefore, provide the user with a sliding control mechanism that effects the agent's tendency to perform autonomous actions versus demonstration requests. Importantly, we note that the overall number of demonstrations required by either approach is less than the teacher-guided method and only a tiny fraction of the overall state space.

## 7. Discussion

In this section, we discuss several promising directions for future work, as well as a number of existing extensions to the presented learning methods.

### 7.1 Evaluation with Non-Technical Users

The presented demonstration learning algorithm provides a fast and intuitive method for programming and adapting the behavior of autonomous agents. We believe that its general representation and classifier-independent approach makes CBA usable for a wide range of applications. One particular application of interest is the use of demonstration learning to enable non-technical users to program autonomous agents. We believe that CBA would be highly suitable for this application as it does not assume that the teacher has any technical knowledge about policy learning, requiring only that the teacher be an expert at the task. The results presented in this article were obtained using only a single teacher, one of the authors. Additional studies could evaluate algorithm usability and performance for a wider user base, and non-programmers in particular.

### 7.2 Representation of Action Choices

Demonstration-based learning provides a natural and intuitive interface for transferring human task knowledge to autonomous agents. However, when operating in rich environments, agents inevitably face situations in which multiple actions are equivalently applicable. For example, an agent that encounters an obstacle directly in its path has the option of moving left or right to avoid it. If the surrounding space is empty, both directions are equally valid for performing the desired task. Human demonstrators faced with a choice of equivalent actions typically do not perform demonstrations consistently, instead selecting among the applicable actions arbitrarily each time the choice is encountered. As a result, training data obtained by the agent lacks consistency, such that identical, or nearly identical, states are associated with different actions. In the presented CBA algorithm, such inconsistent demonstrations would result in a persistent region of low confidence, leading the agent to repeatedly request demonstrations within the inconsistent domain region. We have successfully extended CBA to identify regions of the state space with conflicting demonstrations and represent the choice between multiple actions explicitly within the agent's policy (Chernova & Veloso, 2008a).

### 7.3 Improvement Beyond Teacher Performance

The policy learned by the Confidence-Based Autonomy algorithm is inherently limited by the quality of the demonstrations provided by the human teacher. Assuming that the teacher is an expert at the task, our approach aims to imitate the behavior of the teacher. However, in many domains teacher demonstrations may be suboptimal and limited by human ability. Several demonstration learning approaches have been developed that enable an agent to learn from its own experiences in addition to demonstrations, thereby improving performance beyond the abilities of the teacher (Stolle & Atkeson, 2007; Smart, 2002). Extending the CBA algorithm to include similar capability remains a promising direction for future work. Possible approaches include incorporating a high-level feedback (Argall, Browning, & Veloso, 2007) or reward signal (Thomaz & Breazeal, 2006) from the teacher, as well as filtering noisy or inaccurate demonstrations.

## 7.4 Policy Use After Learning

The CBA algorithm considers learning to be complete once the agent is able to perform the required behavior, repeatedly and correctly, without requesting further demonstrations and requiring corrections. Once policy learning is complete, the standard procedure for the vast majority of policy learning algorithms is to turn off the learning process and freeze the policy. While this approach can also be used with our algorithm, we propose that the continuing use of the Confident Execution component may have long-term benefits beyond policy learning. In particular, the algorithm's ability to identify anomalous states may enable the agent to detect and notify the user of system errors and unexpected input. While further studies are needed to evaluate this use of the algorithm, we believe that such a mechanism would provide a useful safety feature for long-term autonomous operation at a negligible cost of performing the threshold comparison at each timestep.

## 7.5 Richer Interaction

The presented demonstration learning approach relies on a limited form of interaction between the agent and teacher. The agent requests demonstrations from the teacher, while the teacher responds with a single recommended action. While this level of interaction is typical of traditional active learning approaches, it fails to take full advantage of the vast task knowledge that the teacher possesses. We believe that extending the algorithm to include richer interaction abilities could provide a faster and more intuitive training method. Many promising directions for future research exist in this area. For example, developing a domain-independent dialog exchange between the agent and teacher that incorporates clarification questions and high level advice could speed up learning and enable the agent to represent the high level goals of the task. The ability to play back or "rewind" demonstration sequences would additionally enable both teacher and agent to reexamine and reevaluate past learning experiences.

## 7.6 Application to Single-Robot and Multi-Robot Systems

Learning from demonstration techniques have been extensively studied within the robotics community due to their interactive nature and fast learning times. In other work, we have shown the CBA algorithm to be highly effective in learning a variety of single-robot tasks (Chernova & Veloso, 2007, 2008a).

Furthermore, many complex tasks require the collaboration of multiple robots. Up to now, one of the greatest challenges preventing most demonstration learning algorithms from generalizing to multi-robot domains has been the problem of *limited human attention*, the fact that the teacher is not able to pay attention to, and interact with, all robots at the same time. Based on the CBA algorithm, we have developed the first multi-robot demonstration learning system that addresses the limited human attention problem by taking advantage of the fact that the Confident Execution component of CBA prevents the autonomous execution of actions in low-confidence states (Chernova & Veloso, 2008b). Our *flexMLfD* system utilizes individual instances of CBA for each robot, such that each learner acquires a unique set of demonstrations and learns an individual task policy. By preventing autonomous execution in low-confidence states, CBA makes each learner robust to periods of teacher neglect, allowing multiple robots to be taught at the same time.

## 8. Conclusion

In this article we presented Confidence-Based Autonomy, an interactive algorithm for policy learning through demonstration. Using this algorithm, an agent incrementally learns an action policy from demonstrations acquired as it practices the task. The CBA algorithm contains two methods for obtaining demonstrations. The Confident Execution component enables the agent to select demonstrations in real time as it interacts with the environment, using confidence and distance thresholds to target states that are unfamiliar or in which the current policy action is uncertain. The Corrective Demonstration component allows the teacher to additionally perform corrective demonstrations when an incorrect action is selected by the agent. The teacher retroactively provides demonstrations for specific error cases instead of attempting to anticipate errors ahead of time. Combined, these techniques provide a fast and intuitive approach for policy learning, incorporating shared decision making between the learner and the teacher.

Experimentally, we used a complex simulated driving domain to compare five methods of selecting demonstration training data: manual data selection by the teacher, confidence-based selection using a single fixed threshold, confidence-based selection using multiple automatically calculated thresholds, corrective demonstration, and confidence-based selection combined with corrective demonstration. Based on our evaluation, we conclude that all confidence-based methods were able to select more informative demonstrations than the human teacher. Of the single and multiple threshold approaches, the multiple adjustable threshold technique required significantly fewer demonstrations by focusing onto regions of uncertainty and reducing the number of redundant datapoints. The best final policy performance, however, was achieved by the Corrective Demonstration and complete Confidence-Based Autonomy algorithms, both of which achieved a lane preference similar to that of the teacher without any collisions. Together, these demonstration selection algorithms represent the tradeoff between the number of demonstrations and the frequency of undesired actions during training. While Corrective Demonstration required slightly fewer demonstrations to learn the final policy, compared to CBA it resulted in a significant increase in the number of errors made by the agent over the course of the learning process. The CBA algorithm, therefore, provides the best demonstration selection method for domains in which incorrect actions are not desirable during the training process.

## Acknowledgments

This research was partially sponsored by the Department of the Interior, National Business Center under contract no. NBCHD030010 and SRI International under subcontract no. 03-000211, and by BBNT Solutions under subcontract no. 950008572, via prime Air Force contract no. SA-8650-06-C-7606. The views and conclusions contained in this document are those of the authors and should not be interpreted as representing the official policies, either expressed or implied, of any sponsoring institution, the U.S. government or any other entity. Additional thanks to Paul Rybski for making his simulation package available.